\documentclass[11pt,a4paper]{article}
\usepackage[hyperref]{acl2021}
\usepackage{times}  
\usepackage{helvet}  
\usepackage{courier}  
\usepackage{graphicx} 
\usepackage{natbib}  
\usepackage{caption} 
\DeclareCaptionStyle{ruled}{labelfont=normalfont,labelsep=colon,strut=off} 
\usepackage{algorithm}
\usepackage{algorithmic}

\usepackage{newfloat}
\usepackage{listings}
\floatstyle{ruled}
\newfloat{listing}{tb}{lst}{}
\floatname{listing}{Listing}
\usepackage{bibentry}

\usepackage{amsmath,amsfonts,bm}

\def\eqref#1{equation~\ref{#1}}

\def\1{\bm{1}}

\def\rve{{\mathbf{e}}}

\def\rvh{{\mathbf{h}}}

\def\rvs{{\mathbf{s}}}

\def\rvx{{\mathbf{x}}}

\DeclareMathAlphabet{\mathsfit}{\encodingdefault}{\sfdefault}{m}{sl}
\SetMathAlphabet{\mathsfit}{bold}{\encodingdefault}{\sfdefault}{bx}{n}

\usepackage{url}

\usepackage{times}
\usepackage{latexsym}

\usepackage{booktabs}
\usepackage{multirow}
\usepackage{graphicx}

\usepackage{microtype}
\usepackage{amsmath}
\usepackage{amsfonts,amssymb}

\aclfinalcopy 

\usepackage{color}

\title{Response Generation with Context-Aware Prompt Learning}

\author{
    Xiaodong Gu\textsuperscript{\rm 1,2},
    Kang Min Yoo\textsuperscript{\rm 2},
    Sang-Woo Lee\textsuperscript{\rm 2},\\
        \textsuperscript{\rm 1}School of Software, Shanghai Jiao Tong University\\  \textsuperscript{\rm 2}NAVER AI Lab\\
    guxiaodong1987@126.com, kangmin.yoo@gmail.com, sang.woo.lee@navercorp.com\\
}

\date{}

\begin{document}
\maketitle
\begin{abstract}
Pre-trained language models (PLM) have marked a huge leap in neural dialogue modeling. While  PLMs are pre-trained on large-scale text corpora, they are usually fine-tuned on scarce dialogue data with specific domain knowledge and dialogue styles. However, tailoring the language models while fully utilizing prior knowledge in large pre-trained models remains a challenge.
In this paper, we present a novel approach for pre-trained dialogue modeling that casts the dialogue generation problem as a prompt-learning task.
Instead of fine-tuning on limited dialogue data, our approach, DialogPrompt, learns continuous prompt embeddings optimized for dialogue contexts, which appropriately elicit knowledge from the large pre-trained model.
To encourage the model to better utilize the prompt embeddings, the prompt encodings are designed to be dynamically generated based on the input dialogue context.
Experiments on popular conversation datasets show that our approach significantly outperforms the fine-tuning baseline and the generic prompt-learning methods.
Furthermore, human evaluations strongly support the superiority of DialogPrompt in regard to response generation quality.

\end{abstract}

\section{Introduction}
Pre-trained language models (PLMs) such as BERT~\citep{devlin2018bert} and GPT-2~\citep{radford2019gpt2} have achieved remarkable success in various natural language processing tasks~\citep{sun2019ernie}. As such, there is a growing trend of using pre-trained language models for conversation modeling~\citep{budzianowski2019hello,zhang2019dialogpt,feng2021language}. For example, \citet{zhang2019dialogpt} proposed DialoGPT, a dialogue generation model that trains an extended GPT-2~\citep{radford2019gpt2} on large dialogue corpus.
\citet{feng2021language} further explore the usage of DialoGPT for dialogue summarization. 
These pre-trained dialogue models are often pre-trained on large text corpora and fine-tuned on smaller dialogue datasets~\cite{zhang2019dialogpt}.

One limitation of PLM-based dialogue modeling, and even for other PLM tasks, is the trade-off between pre-training and fine-tuning~\citep{ben2021pada}. That is, the task-specific data used for fine-tuning is usually scarce and costly. As such, the reusability of prior knowledge learned in the pre-training phase can be limited during fine-tuning, hence some dialogue models are simply trained from scratch on the limited task-specific data. 

Consequently, recent works have resorted to prompt learning, a lightweight alternative to fine-tuning. Prompt learning keeps the PLM parameters frozen but optimizes only a small portion of task-specific prompts or related modules~\citep{liu2021survey,shin2020autoprompt,liu2021ptuning,li2021prefix}. For example, \citet{liu2021ptuning} propose \emph{p-tuning}, which preprends trainable prompt tokens to the input of a PLM. The trainable prompt embeddings are optimized while the PLM parameters are kept frozen. Prompt learning allows few-shot or nearly zero-shot learning for pre-trained models in new tasks with little or unlabeled data and it has been demonstrated to be substantially effective over fine-tuning in many tasks~\cite{liu2021survey,qin2021learning}. 

However, applying prompt learning directly to conversation modeling is challenging. The general prompt-learning models assign universal prompt tokens to all inputs in the same task~\cite{liu2021ptuning}. For example, prompts used for sentiment analysis share the same embeddings that are inferred from the training data~\cite{liu2021ptuning}. In contrast, conversations are context-sensitive. Dialogue responses are affected by contextual information, such as the topic of discussion, pre-dialogue context, and participant personalities. ``Blanket'' prompts can restrict the expressiveness of prompt learning due to the lack of context-awareness, leading to sub-optimal performance in response generation.

In this work, we present DialogPrompt, a novel prompt-based paradigm for response generation on top of large pre-trained language models. 
DialogPrompt prepends a sequence of prompt tokens to each dialogue context for eliciting response from large pre-trained language models. In order to construct context-aware prompts, we propose a dynamic prompt encoder on top of the Transformer \citep{vaswani2017attention}. The prompt tokens are initially encoded conditionally on the dialogue context. The resulting prompt encoding is then taken as the initial hidden state of the large PLM to generate responses. Compared to fine-tuning, DialogPrompt is encouraged to search proper prompts which controls the large PLMs into producing higher-quality responses directly.

We evaluate DialogPrompt on popular multi-turn conversation datasets such as DailyDialog and MultiWOZ. Results show that DialogPrompt outperforms fine-tuning counterparts and other prompt tuning methods in terms of automated evaluation measures and the average length of generated responses. Human evaluation supports the superiority of our approach in generating informative and knowledgeable responses.

Our contributions are summarized as follows:
\begin{itemize}
    \item To the best of our knowledge, we are the first to propose prompt-based learning for general dialogue generation. Our approach can better reuse knowledge from existing large-scale PLMs and produce more knowledgeable responses.
    \item We design a novel dynamic prompt encoder for encouraging context-aware prompt learning.
    \item We extensively evaluated our approach on popular multi-turn conversation datasets and demonstrated the superiority of our approach in terms of quantitative automatic evaluations and qualitative human evaluations.
\end{itemize}

\section{Related Work}
This work is closely related to (1) pre-trained models for conversations, and (2) prompt learning for pre-trained language models.

\noindent\textbf{Pre-trained Models for Dialogue Generation}. 
Recently, an emerging trend in dialogue generation explores the adaptation of large pre-trained language models on dialogue corpora~\cite{golovanov2019transfertransfo,zhang2019dialogpt}.
For example, \citet{golovanov2019transfertransfo} studied how pre-trained architectures can be adapted for natural language generation, comparing a number
of architectural and training schemes. 
The state-of-the-art DialoGPT~\cite{zhang2019dialogpt} pre-trains a GPT-2 model on large-scale conversation datasets and achieves a giant leap in performance against traditional conversation models. 

Another line of work related to exploiting the use of pre-trained models for task-oriented dialogues.
For example, \citet{budzianowski2019hello} proposed a task-oriented dialogue model that operates solely on text input. Their model is built on top of the TransferTransfo framework~\cite{golovanov2019transfertransfo} that effectively bypasses explicit policy and language generation modules. TOD-BERT proposed by \citet{wu2020todbert} bridges the difference of general text and task-oriented dialogue by unifying nine human-human and multi-turn task-oriented dialogue datasets for language modeling. The model also incorporates user and system tokens into the masked language modeling and proposes a contrastive objective function to simulate the response selection task.

Compared to these related works which directly fine-tune the dialogue model based on a pre-trained model, DialogPrompt is a novel paradigm for pre-trained dialogue models which elicits knowledge from PLMs directly through minimal optimizing of prompt tokens.

\noindent\textbf{Prompt Learning for Pre-trained Language Models}. 
There is a growing trend of automatically finding prompts to adapt pre-trained language models to downstream tasks~\cite{shin2020autoprompt,li2021prefix,liu2021ptuning}. For example,
\citet{shin2020autoprompt} proposed AutoPrompt which automatically optimizes prompts using a gradient signal. Unlike our method, AutoPrompt searches for hard prompts, thus it may be less versatile than the continuous methods.
Instead, \citet{liu2021ptuning} proposed a continuous prompt tuning model named p-tuning. p-tuning optimizes fill-in-the-blank prompts in a continuous space, tested on GPT-2 and BERT models.
A similar idea was proposed by Li and Liang~\citep{li2021prefix} who considered the tuning of prompts using a textual prefix. Specifically, they prepended a few task-specific ``soft tokens'' (prefix) to the source text and tuned the hidden states of only these tokens (at all Transformer layers). 
Similarly, \citet{lester2021power} prepended a sequence of prompt tokens to the source text, but only the word embeddings of these tokens are optimized.
\citet{qin2021learning} proposed prompt-based learning on relation extraction tasks using data-dependent mixtures of prompt templates and parameters.

Our method differs from existing prompt-based tuning methods in that we propose a novel context-aware prompt tuning mechanism that can optimize prompt encodings conditioned on dialogue contexts. Our work also differs from a very recent work by Zheng et al. which explores prompt-based learning for grounded dialogue generation~\cite{zheng2021exploring}.

\section{Approach}
\label{sec:approch}
\begin{figure*} [!tb]
		\centering 
		\includegraphics[width=6in]{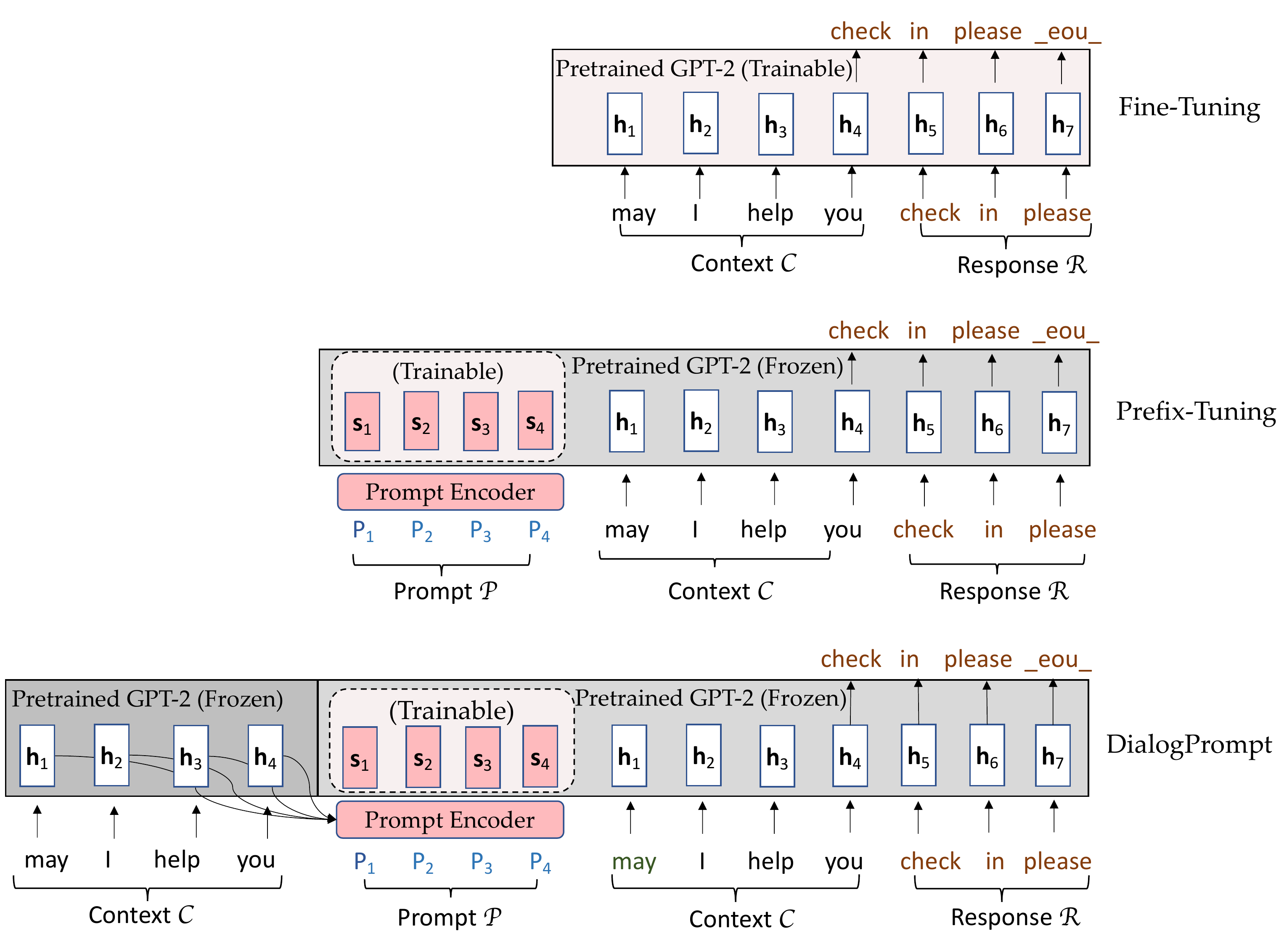} 
		\caption{Overview of fine-tuning and (context-aware) prompt-tuning for response generation.}
		\label{fig:arch}
	\end{figure*}

In this section, we present the implementation of DialogPrompt. The overall framework is shown in Figure~\ref{fig:arch}.
First, we present the standard autoregressive pre-trained dialogue model as the backbone. Then, we will introduce a naive idea of applying prompt learning for dialogue modeling, followed by our novel context-aware prompt tuning model for dialogues.

\subsection{Response Generation via Autoregressive Transformer Models} 
\label{ss:finetuning}
Let $\mathcal{D}$ = [$x_1,\ldots,x_N$] denote a dialogue of $N$ words which is comprised of a context $\mathcal{C}$ = [$x_1,\ldots,x_m$] followed by a response $\mathcal{R}$ = [$x_{m+1},\ldots,x_N$]. The goal of response generation is to produce the response given the dialogue context, namely, estimating the conditional probability of $p(\mathcal{R}|\mathcal{C})$. 

As shown in Figure~\ref{fig:arch}~(top), an autoregressive Transformer model such as GPT-2~\citep{radford2019gpt2} solves this problem by sequentially estimating the probability of each word in the target response conditioned on historical words in the dialogue:
\begin{equation}
    p(\mathcal{R}|\mathcal{C}) = \prod_{i=m+1}^N p_\phi(x_i|x_1,\ldots,x_{i-1})
\end{equation}
where $\phi$ denotes the trainable parameters of the autoregressive Transformer model.

In detail, let $\rve(\mathcal{D}$) = [$\rve(x_1),\ldots, \rve(x_N)$] be the embeddings of the dialogue tokens. These input embeddings are then fed into the pre-trained Transformer to obtain the contextual representations $\mathbf{H}$ = [$\rvh_1,\ldots,\rvh_N$], where each $\rvh_i$ is a function of $\rvx_i$ and the past representations of its left context:
\begin{equation}
   \label{eq:lm}
    \rvh_i = \text{Transformer}_\text{pre}(\rve(x_i), \rvh_{<i})
\end{equation}
Then, each $\rvh_i$ is used to compute the distribution for the next token: $p(x_{i+1}| \rvh_{\leq i})$ = softmax($\mathbf{W} \rvh_i$) and $\mathbf{W}$ is a pre-trained matrix that maps $\rvh_i$ to logits over the vocabulary.

In the conventional fine-tuning framework, we initialize with the pre-trained parameters $\phi_0$ and perform gradient descend on the following objective:
\begin{equation}
  \label{eq:gptloss}
   \begin{split}
    \mathcal{L}_\mathrm{dec} (\phi|\mathcal{R}, \mathcal{C}, \phi_0) & =  -\text{log}~p_\phi(\mathcal{R}|\mathcal{C}) 
    \\
    & = -\sum_{i=m+1}^N\text{log}~p_\phi(x_i|\rvh_{<i})
    \end{split}
\end{equation}
where $p_\phi$ denotes the trainable language model.

\subsection{Prompt Learning for Conversations}	

Based on intuition from prompting~\citep{liu2021ptuning,li2021prefix}, we believe that having a proper prompt of context can adapt the large pre-trained language model to the conversation domain without re-training all its parameters from scratch~\citep{ben2021pada}.

One intuitive baseline approach can be proposed to simply adopt previous work in prompt learning (e.g., prefix-tuning on GPT-2~\cite{li2021prefix}) for conversations. More specifically, we can prepend a prompt utterance of $k$ tokens $\mathcal{P}$ = [$p_{1},\ldots,p_{k}$] to each dialogue context to obtain $\tilde{\mathcal D}$ = [$\mathcal{P}$; $\mathcal{C}$; $\mathcal{R}$], as shown in Figure~\ref{fig:arch}~(middle).
A fully connected prompt encoder can be designed to transform the prompt utterance into a sequence of hidden states, namely, 
\begin{equation}
    \rvs_{1}, \ldots, \rvs_{k} = f_\theta(\rve(p_1),\ldots,\rve(p_k))
\end{equation}
where $f_\theta$ denotes the fully connected neural network and $\theta$ denotes the trainable parameters.

We follow the same recurrence relation in Equation~\ref{eq:lm}, except that the hidden states $s_{1:k}$ of the prompt utterance are taken as past hidden states of the Transformer:
\begin{equation}
\label{eq:prompt_hids}
    \rvh_i = \text{Transformer}_\text{pre}(x_i, \rvs_{1:k}, \rvh_{<i})
\end{equation}
Now the transformer hidden states depend on the prompt encodings, because the prompt utterance is always located to the left of the context and hence affect hidden states of the context in the autoregressive Transformer.

The training objective is to only optimize the prompt parameters $\theta$ while keeping the pre-trained Transformer parameters $\phi$ frozen, namely,
\begin{equation}
    \mathcal{L}_\mathrm{prompt}(\theta|\mathcal{R}, \mathcal{C}, \phi) = - \sum_{i=m+1}^N \mathrm{log} p_\phi(x_i|\rvs_{1:k}, \rvh_{<i})
\end{equation}
where $\theta$ denotes the only trainable parameters of the fully connected prompt encoder; $\phi$ represents the frozen parameters of the pre-trained language model.

\subsection{Dynamic Prompt Learning for Context-Aware Prompt Adaptation}
In the previous model, the encoding of the prompt utterance {$\rvs_{1}, \ldots, \rvs_{k}$} is independent of dialog context $\mathcal{C}$. That means, prompts for all conversations share the same encoding. However, the latent space of dialogue context is more complicated and difficult to be represented with such a unified encoding. 
Intuitively, the context can influence the encoding of prompt by guiding what to extract from the PLM. We want to find a prompt encoding that steers the LM to the current context. 

Extending this intuition beyond generating a unified prompt encoding, we propose a dynamic prompt encoder, as shown in Figure~\ref{fig:arch}~(bottom). Given the dialogue context $\mathcal{C}$=[$x_1,\ldots,x_m$] with a prompt utterance $\mathcal{P}$=[$p_1,\ldots,p_k$], the prompt encodings [$\tilde{\rvs}_1,\ldots,\tilde{\rvs}_k$] are dynamically generated conditioning on the context using another autoregressive Transformer: 
   \begin{equation}
       \tilde{\rvs}_i = \text{Transformer}_\text{prompt}(p_i, \rvh_1,\ldots, \rvh_m, \tilde{\rvs}_{<i})
   \end{equation}
where $\rvh_1,...,\rvh_m$ are computed using Equation~\ref{eq:lm} and are taken as past hidden states for the new Transformer to generate the prompt encodings.

Now we update the hidden states of the pre-trained Transformer based on the new prompt encodings: 
\begin{equation}
\label{eq:prompt_hids}
    \tilde{\rvh}_i = \text{Transformer}_\text{pre}(x_i, \tilde{\rvs}_1,\ldots,\tilde{\rvs}_k, \tilde{\rvh}_{<i})       
\end{equation}

The final hidden states [$\tilde{\rvs}_1,\ldots,\tilde{\rvs_k},\tilde{\rvh_1},\ldots,\tilde{\rvh_i}$] are taken as input the the pre-trained language model to generate the response.
Our training objective now becomes to minimize the following loss function:
\begin{equation}
\begin{split}
    &\mathcal{L}_\mathrm{dyn\_prompt}(\theta|\mathcal{R},\mathcal{C},\phi) \\
     &= - \sum_{i=m+1}^N \mathrm{log}\;p_{\phi,\theta}(x_i|\tilde{\rvs}_{1:k},\tilde{\rvh}_{<i})
\end{split}
\end{equation}
where $\theta$ represents the parameters for the prompt encoder; $\phi$ denotes the frozen parameters of the pre-trained language model.

\section{Experimental Setup}

\subsection{Dataset}
We evaluate all models on two popular response generation datasets, namely, DailyDialog\footnote{http://yanran.li/dailydialog} and MultiWOZ\footnote{https://github.com/budzianowski/multiwoz}. 
Table~\ref{table:dataset_overview} shows the statistics of these datasets.
\begin{table}
\centering
\begin{tabular}{lcc}
\toprule
\textbf{Dataset} & \bf DailyDialog &  \bf MultiWOZ \\
\midrule
dialogues   & 13,118   &  8,438 \\ 
train samples   & 76,052 & 106,794 \\
valid samples   & 7,069  & 12,902 \\
test samples  & 6,740 & 12,914 \\ 
\bottomrule
\end{tabular}
\caption{Overview of the datasets.}
\label{table:dataset_overview}
\end{table}
The DailyDialog is a manually labeled multi-turn dialogue dataset that contains daily conversations in English. As was originally designed for English learners, DailyDialog has a more chit-chat style compared to other datasets.
The MultiWOZ~\citep{budzianowski2018multiwoz} is a fully-labeled collection of human-human written conversations spanning over multiple domains and topics such as attraction, hotel, hospital, police, restaurant, train, and taxi. Compared to DailyDialog, MultiWOZ is more challenging due to the diverse domains and language styles.

\subsection{Implementation Details}
We used GPT-2~\citep{radford2019gpt2} as the backbone PLM for all models. GPT-2 has been widely employed for generating dialogues~\cite{zhang2019dialogpt}. We did not take the more advanced GPT-3 due to the restriction of our computational resources. Besides, GPT-3 relies heavily on super large models and training corpora, so the effect of prompt learning could be overwhelmed.
Our implementation was based on the Huggingface Transformer repository~\citep{Wolf2019HuggingFacesTS}. 
For the sake of computational efficiency, we limit each context to have at most 4 utterances, with each containing less than 20 words. The batch size for all models was set to 32.
In the generation phase, we used the top-1 sampling for response decoding. 
The hyperparameters we tuned include the prompt size and learning rate. We search for the best hyperparameters using NAVER Smart Machine Learning (NSML)~\cite{sung2017nsml,NSML,park2019visualhypertuner}. The prompt size was empirically set to 5 and 20 for DailyDialog and MultiWOZ, respectively. 
All models were optimized using the AdamW~\citep{loshchilov2018adamw} optimizer using initial learning rates of 1$e$-3 and 5$e$-5 for DailyDialog and MultiWOZ, respectively. We used a linear learning rate scheduler with 5,000 warm-up steps. 
We trained all models on a Linux server with Ubuntu 16.04 and a GPU of Nvidia Tesla V100.
The training processes were early stopped when there was no progress on the validation loss. Then, the corresponding checkpoint is used to evaluate the performance on the test set.
We run each experiment for five times and reported the average scores.

\subsection{Baseline Models}
We compare our approach with popular fine-tuning and prompt learning methods, namely, 
\\
(i) \textbf{Fine-Tuning}: the default training method for adapting pre-trained models to conversations. As we use GPT-2 as our backbone model, we implement this baseline model by directly fine-tuning GPT-2~\citep{radford2019gpt2} on the conversation datasets; 
(ii) \textbf{P-Tuning}~\citep{liu2021ptuning}: a well-known prompt learning method which searches continuous prompts by adding prompt tokens. In our implementation, we modify the GPT-2 input by prepending prompt tokens before each utterance in the context. Specifically, we construct the following template $\{[P_{0:k}],u_1,[P_{k+1:2k}],u_2,\ldots\}$ for GPT-2, where $k$ denotes the number of prompt tokens for each utterance. We empirically set $k$ to 3 in our experiments.
(iii) \textbf{Prefix-Tuning}~\citep{li2021prefix}: which prepends a prefix of prompt tokens before the source sequence and optimizes hidden states of all Transformer layers for the prompt tokens. Prefix-tuning is a similar approach to our DialogPrompt except that it uses a unified prompt encoding for all input contexts. Therefore, we implemented the common modules using the same configuration as in DialogPrompt; and 
(iv) \textbf{Soft-Prompt-Tuning}~\citep{lester2021power}: is similar to prefix-tuning in architecture. The difference between two methods is in that softprompt-tuning only optimizes the embeddings of prompt tokens, while prefix-tuning optimizes parameters of all hidden layers.

\begin{table*}[!htb]
\centering
\begin{tabular}{lccc@{}cc}
\toprule
\bf Model & \bf \;BLEU\; & \bf NIST\; & \bf METEOR\;\; & \bf ROUGE-L & \bf Length\\
\midrule
Fine-Tuning & 12.65	& 17.55	& 7.04	& 8.01 & 15.60\\
P-Tuning  & 10.36  & 12.54  & 5.09  &  5.91 & 15.28\\
SoftPrompt-Tuning & 10.25 & 13.22  & 5.20  &  6.20 & 14.46 \\ 
Prefix-Tuning  & 12.29  &  16.64  &  6.55  & 7.55 &  15.49\\ 
\midrule
DialogPrompt (ours)\;  & 13.94 & 19.07 & 7.61 & 8.51 & 16.90\\ 
\bottomrule
\end{tabular}
\caption{Comparison between DialogPrompt and baseline models on the DailyDialog dataset.}
\label{table:results:dailydial}
\end{table*}

\begin{table*}[!htb]
\centering
\begin{tabular}{lccc@{}cc}
\toprule
\bf Model & \bf \;BLEU\; & \bf NIST\; & \bf METEOR\;\; & \bf ROUGE-L & \bf Length \\
\midrule
Fine-Tuning & 20.31 & 47.82 & 16.54 & 17.34 & 18.70\\ 
P-Tuning  & 16.66  &  24.60 & 7.57  & 8.93  & 18.03\\ 
SoftPrompt-Tuning & 16.64  & 24.40  &  7.18 & 8.78 & 18.08 \\ 
Prefix-Tuning  & 20.24  &  48.68  &  16.25  & 17.24 & 18.21 \\ 
\midrule
DialogPrompt (ours)\; & 20.96 & 52.94 & 17.62 & 18.22 & 18.91\\
\bottomrule
\end{tabular}
\caption{Comparison between DialogPrompt and baseline models on the MultiWOZ dataset.}
\label{table:results:multiwoz}
\end{table*}

\subsection{Evaluation Metrics}
We evaluate all models using five commonly used metrics in NLG, namely, BLEU~\citep{papineni2002bleu}, NIST~\citep{doddington2002nist}, METEOR~\citep{lavie2007meteor}, ROUGE-L~\citep{lin2004rouge}, and the average length of generated responses.

BLEU evaluates how many n-grams in the generated response match those in the human reference.
We report the average of BLEU 1-4 scores in our experiments using the NLTK toolkit\footnote{https://www.nltk.org/modules/nltk/translate/bleu\_score.html}. 
NIST~\citep{doddington2002nist} is similar to BLEU but assigns different weights to n-gram matches according to their information gain. We use the implementation in the NLTK toolkit\footnote{https://www.nltk.org/modules/nltk/translate/nist\_score.html}.
METEOR~\citep{lavie2007meteor} is based on unigram matching (surface forms, stemmed forms, and meanings) between the generated response and human reference. 
ROUGE-L~\citep{lin2004rouge} measures the longest common subsequence (LCS) between the generated response and human reference. 
Finally, the average length of generated responses is also a critical metric to measure the quality of generated responses~\cite{gu2019dialogwae,zhang2019dialogpt}. Research has shown that dialogue models can produce safe responses that are usually short and uninformative (e.g., I do not know)~\cite{gu2019dialogwae}. We simply average the length of generated responses for all test examples.

\section{Evaluation Results}

\subsection{Automatic Evaluation}

Table~\ref{table:results:dailydial} and \ref{table:results:multiwoz} shows the performance of each method on the two datasets respectively.
Broadly, DialogPrompt achieves the best performance across all automatic metrics. Compared to the fine-tuning baseline, DialogPrompt is superior across all metrics with a large margin. For example, the BLEU score is increased by 10\% on the DailyDialog dataset. Such improvement is consistent on both datasets, affirming the superiority of the prompt based pre-trained dialogue model. 
This indicates that our approach by eliciting knowledge from pre-trained GPT-2 is more effective than the fine-tuning counterparts. 

The prefix-tuning baseline model achieves similar performance to that of fine-tuning, which indicates that simply applying previous prompt learning methods to dialogues does not bring better performance to PLM-based dialogue models. 

Among the three prompt learning models (i.e., p-tuning, softprompt-tuning, and prefix-tuning), the prefix-tuning achieves the best performance on both datasets. This is probably because the prefix-tuning optimizes activations of all layers, which is more compatible to autoregressive Transformers such as GPT-2. Besides, the prefix-tuning optimizes more parameters compared to simply optimizing prompt embeddings~\cite{li2021prefix,liu2021ptuning}.

The improvement of DialogPrompt on the MultiWOZ dataset is less significant in terms of BLEU and average response length compared to that on the DailyDialog dataset. This is probably because the MultiWOZ dataset was originally prepared for task-oriented dialogues and contains specific domain knowledge. Such task-specific knowledge might not contained in the pre-trained GPT-2. Another possible reason for this phenomenon is that MultiWOZ has a larger amount of data, which brings more efficacy for fine-tuning methods which are usually data hungry.


Overall, these results show that DialogPrompt can utilize pre-trained language models more effectively than general prompt learning methods.

\subsection{Ablation Study}

\begin{figure} [tb]
	\centering 
	\includegraphics[width=2.5in]{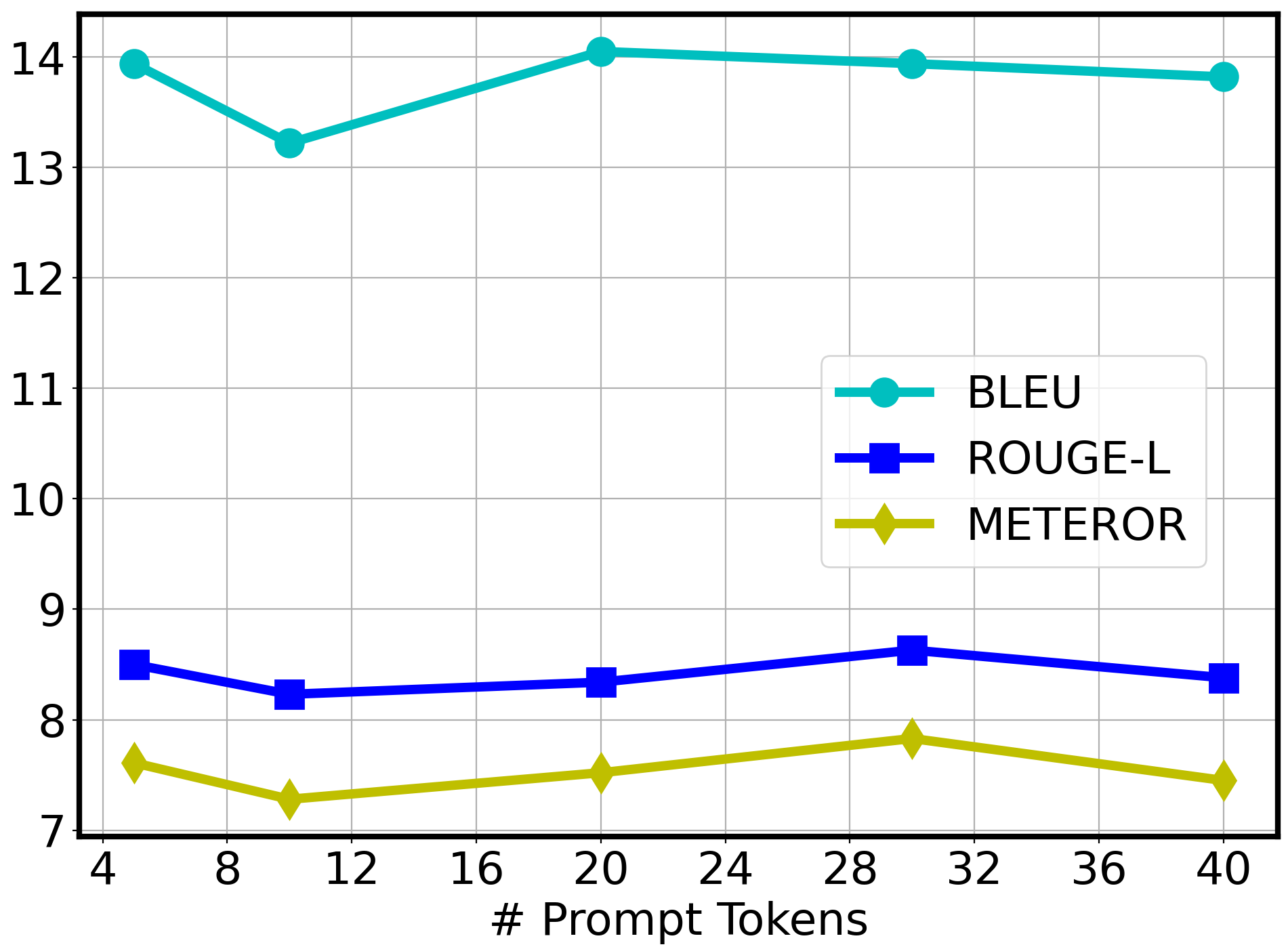} 
	\caption{Effects of prompt size on the DailyDialog dataset. For ease of rendering, we omit the results in terms of NIST.}
	\label{fig:ablation:prompt}
\end{figure}

\begin{table*}[!htb]
\centering
\begin{tabular}{lccc@{}cc}
\toprule
\bf Model & \bf \;BLEU\; & \bf NIST\; & \bf METEOR\;\; & \bf ROUGE-L & \bf Length \\
\midrule
Fine-Tuning (standard) & 12.65 & 17.55 & 7.04 & 8.01 & 15.60\\ 
DialogPrompt (standard)  & 13.94 & 19.07 & 7.61 & 8.51 & 16.90 \\ 
\midrule
Fine-Tuning (medium) &  14.72   & 22.27 &  8.75  & 10.16 & 15.93  \\ 
DialogPrompt (medium)  & 14.98 &  22.79 &  9.44 & 10.75  & 16.17 \\ 
\midrule
Fine-Tuning (large) & 15.77 & 23.84  & 10.59  & 11.41  &  15.97\\
DialogPrompt (large)\; & 16.60 & 26.61 &  11.39  &  12.35   & 16.76 \\
\bottomrule
\end{tabular}
\caption{Comparison between DialogPrompt and baseline models with different model sizes in the DailyDialog dataset.}
\label{table:ablation:size}
\end{table*}

One of the key hyperparameters of our approach is the prompt size, namely, the number of prompt tokens for each context. We conducted an ablation study to assess the sensitivity of prompt size to the performance.
We trained DialogPrompt on the DailyDialog dataset with various numbers of prompt tokens in the prompt utterance.
As shown in Figure~\ref{fig:ablation:prompt}, the number of prompt tokens has little effect on the performance. The model can achieve satisfactory performance with only 5 prompt tokens. Increasing the prompt size does not bring significant improvement to the performance. 
Considering the balance between performance and complexity, the optimal number of prompts on the DailyDialog dataset is around 5. 

We also conducted an ablation study on the model size of the pre-trained models.
We trained DialogPrompt with different GPT sizes, namely, standard (L=12, H=12, D=768), medium (L=24, H=16, D=1024), and large (L=36, H=20, D=1280). 
Results show that DialogPrompt outperforms the fine-tuning counterpart with all three model sizes. That means that our method is effective on different scales of backbone pre-trained models. However, as the model size increases, the improvement of our method becomes less significant, which indicates that DialogPrompt is more effective in smaller pre-trained models. We conjecture that larger pre-trained models contain massive parameters and were pre-trained with enormous data which may overwhelm the efficacy of prompt tuning.
Nevertheless, we found that the model size has a positive correlation to the performance: larger GPT-2 backbones tend to achieve better performance.



\begin{table*} [!htb]
\centering
\small
\begin{tabular}{lccccccccc}
\toprule 
\multirow{2}{*}{\textbf{Comparison}} &
    \multicolumn{3}{c}{{Coherence}} & \multicolumn{3}{c}{{Informativeness}} & \multicolumn{3}{c}{{Fluency}} \\
\cline{2-10}
 & \textbf{Win} & \textbf{Tie} & \textbf{Loss} &
    \textbf{Win} & \textbf{Tie} & \textbf{Loss} &
    \textbf{Win} & \textbf{Tie} & \textbf{Loss} \\
\midrule
Ours \textit{vs.} Fine-Tuning &  58.13\% & 18.79\% & 23.07\%  & 56.43\% & 23.16\%  & 20.41\% & 57.42\% & 20.45\% & 22.13\% \\
Ours \textit{vs.} P-Tuning & 60.60\% & 19.59\% & 19.82\%  & 58.82\%  & 24.15\%  & 17.03\%  & 59.38\% & 21.15\% & 19.48\% \\
Ours \textit{vs.} Prefix-Tuning & 58.13\% & 18.79\% & 23.07\% & 55.89\% & 22.94\% & 21.16\% & 57.36\% & 20.43\% & 22.21\% \\
\bottomrule
\end{tabular}
\caption{Human evaluation on the test set of DailyDialog. All results have statistical significance of $p<0.001$.}
\label{table:results:human}
\end{table*}

\subsection{Human Evaluation}

To further verify the effectiveness of DialogPrompt, we conducted a human evaluation on the Amazon Mechanical Turk platform. We chose DailyDialog as the evaluation corpus since it is in the style of daily chit-chats and can be easier for annotators to rate their preference.
We randomly sampled 200 dialogues from the test set of DailyDialog. For each of the samples, we present the dialogue context, followed by a pair of responses from our model or a baseline model (without order), to three different workers. Each worker was asked to evaluate the responses in terms of the three criteria, namely, coherence, informativeness, and fluency. Coherence measures how relevant the generated responses are to the context. Informativeness measures how well the generated response includes non-trivial information. Fluency measures how well the generated responses are human readable. Finally, the workers blindly rated their preference using a 3-point Likert scale: ``win'' (ours is better), ``loss'' (the other is better) and ``tie'' (equally good or bad). 

The overall results is summarized in Table~\ref{table:results:human}.
A strong preference can be observed for DialogPrompt over baseline models in terms of all metrics.
Interestingly, fine-tuning and prefix-tuning receive similar preferences, which is consistent to the results in the automatic evaluation. 

\begin{table*} [!htb]
\centering
\begin{tabular}{rl}
\toprule

\multirow{2}{*}{\textbf{Context:}} & A: Good evening. What'll you have? \\ 
                             & B: I'd like a beer. What kind of beer do you have? \\
                             \hline
                             \hline
  \bf fine-tuning: & I want a light to medium beer. \\ 
  \bf p-tuning: &  I will be glad to help you out once your stay here gets a little hot.\\ 
  & Thank you. \\
  \bf prefix-tuning:  & Pardon? \\ 
  \bf DialogPrompt: & We have the {\bf Imperial Porter}. It's our {\bf standard} beer. \\
  \bf Human: & Draft beer and cans beer. Which do you prefer?  \\
\bottomrule
\end{tabular}

\quad

\quad

\begin{tabular}{rl}
\toprule
\multirow{4}{*}{\textbf{Context:}} & A: how long will it take us to drive to London? \\
       & B: I think it’ s a distance of 180 kilometers from here to London, \\
       & \; \; so it should be a two-hour drive on the motorway. \\
       & A: that’ s unless there is a traffic jam. It could take three hours.\\
\hline\hline
\textbf{fine-tuning}: & I think it will be difficult, but I think we’ ll enjoy ourselves.\\ 
\textbf{p-tuning}: & I’ t think I want to stay here. \\
\textbf{prefix-tuning}:  & it can be dangerous. You could be killed. \\ 
\textbf{DialogPrompt}:  &  I think it would be super \textbf{time-consuming}. It would also be bad \\
 &  for \textbf{business}. There could be a \textbf{car accident}, especially if \\
 &  something goes wrong.\\
\textbf{Human}: & You’ re right. We will be able to travel at high speeds at the \\
 & beginning and end.\\
\bottomrule
\end{tabular}
\caption{Sample conversations from multiple models with human reference.}
\label{table:case}
\end{table*}

\subsection{Case Study}
  \label{ss:results:case}

Table~\ref{table:case} presents two sample responses generated by different models in the DailyDialog dataset.
As the samples indicate, DialogPrompt generates more coherent responses than other models, which is consistent with the results of the automatic and human evaluation.
According to the first sample, instead of giving a safe response, DialogBERT produces a more contentful response. For example, the response contains specific information such as ``\emph{imperial porter}'' and ``\emph{standard}''. This is consistent with the results of informativeness in the human study. 
The second sample shows more clear strength of DialogPrompt which produces the longest response among all the results. Besides, the response generated by DialogPrompt contains knowledgeable, human-like keywords, such as `time-consuming', `business', and `car accident'. This is presumably because DialogPrompt reuses much knowledge from pre-trained models which have already seen large amounts of domain-specific data. 
DialogPrompt also shows a more fluent responses than baseline models. For example, the baseline model p-tuning generates a response that contains grammar errors such as `I't think' and `gets a little hot', while our approach generates error-free and human-like responses. 

Our observations suggest that DialogPrompt is better in modeling multi-turn conversations than fine-tuning counterparts and naive adaptations of existing prompt learning models.

\section{Discussions}

\subsection{Why does DialogPrompt work better than fine-tuning?}
One possible reason for the improvement is the intensified learning of reusing knowledge from PLMs by freezing the autoregressive decoder. The fine-tuning baseline model attributes all variations of data to the autoregressive decoder. A sufficiently high-capacity autoregressive decoder can model the conditional density directly, ignoring the relations p($\mathcal{R}|\mathcal{C}$) between contexts and responses~\cite{mccarthy2020addressing}. 
By freezing the decoder, our model restricts the optimization to the local prompt Transformer. Hence, it focuses more on how to reuse knowledge from the pre-trained model instead of decoding the target response autoregressively.


\subsection{Threats to Validity}
Our model is built on top of the GPT-2 model. Although GPT-2 is one of the most typical pre-trained models and has been shown to be effective in response generation~\cite{zhang2019dialogpt}, it still remains to be verified whether or not the proposed prompt model is applicable to other pre-trained models. We leave prompt learning on other pre-trained models for future directions.

\section{Conclusion}
In this paper, we propose DialogPrompt, a novel prompt based response generation model.
DialogPrompt prepends a prompt utterance to the dialogue context and only optimizes the prompt encoder. In order to adapt to different contexts, we propose a dynamic prompt encoder that updates prompt activation based on the hidden states of context before response generation. 
Results on two popular conversation datasets, namely, DailyDialog and MultiWOZ show that DialogPrompt significantly outperforms fine-tuning counterparts and other prompt based models on both automatic and human evaluations.
In the future, we will investigate prompt-based dialogue modeling based on more pre-trained language models.

\section*{Acknowledgments}
The author would thank Jung-Woo Ha at NAVER AI Lab for his support and valuable comments on this project. This work was done when the first author was visiting NAVER AI Lab.

\bibliography{references}
\bibliographystyle{acl_natbib}



\end{document}